\documentclass[letterpaper, 10 pt, conference]{ieeeconf}  

\IEEEoverridecommandlockouts                              

\usepackage{amssymb}
\usepackage{pifont}
\usepackage{multirow}
\usepackage{graphicx}
\usepackage{float}  
\usepackage{amsmath}
\usepackage{amssymb}
\usepackage{booktabs}
\usepackage[inkscapeformat=pdf]{svg}
\usepackage{colortbl}
\usepackage{svg}
\usepackage{pgfplots}
\usepackage{graphicx}
\usepackage{subcaption}
\usepackage{makecell}
\usepackage{hhline}

\title{\LARGE \bf
M$^3$CAD: Towards Generic Cooperative Autonomous Driving Benchmark
}

\author{Morui Zhu$^{1}$, Yongqi Zhu$^{1}$, Yihao Zhu$^{1}$, Qi Chen$^{2}$, Deyuan Qu$^{2}$, Song Fu$^{1}$, Qing Yang$^{1\dagger}$
\thanks{$^{1}$University of North Texas}%
\thanks{$^{2}$Toyota InfoTech Labs}%
\thanks{$^{\dagger}$For correspondence and questions: qing.yang@unt.edu}
}

\begin{document}

\maketitle
\thispagestyle{empty}
\pagestyle{empty}

\begin{abstract}

We introduce M$^3$CAD, a comprehensive benchmark designed to advance research in generic cooperative autonomous driving. M$^3$CAD comprises 204 sequences with 30,000 frames. Each sequence includes data from multiple vehicles and different types of sensors, e.g., LiDAR point clouds, RGB images, and GPS/IMU, supporting a variety of autonomous driving tasks, including object detection and tracking, mapping, motion forecasting, occupancy prediction, and path planning. This rich multimodal setup enables M$^3$CAD to support both single-vehicle and multi-vehicle cooperative autonomous driving research. 
To the best of our knowledge, M$^3$CAD is the most complete benchmark specifically designed for cooperative, multi-task autonomous driving research.
To test its effectiveness, we use M$^3$CAD to evaluate both state-of-the-art single-vehicle and cooperative driving solutions, setting baseline performance results.
Since most existing cooperative perception methods focus on merging features but often ignore network bandwidth requirements, we propose a new multi-level fusion approach which adaptively balances communication efficiency and perception accuracy based on the current network conditions.
We release M$^3$CAD, along with the baseline models and evaluation results, to support the development of robust cooperative autonomous driving systems. All resources will be made publicly available on our project webpage.

\end{abstract}

\section{INTRODUCTION}

Cooperative autonomous driving (CAD) refers to the paradigm where multiple autonomous vehicles communicate and coordinate with each other to enhance driving efficiency and safety.
To advance research in this domain, there is an urgent need for a comprehensive benchmark that enables evaluation and comparison of CAD algorithms and solutions.
To fill this gap, we introduce a benchmark with the following key features:
(1) It provides scenarios involving multiple vehicles, focusing on cases where they can collaborate with each other. 
(2) It supports research on a variety of cooperative driving tasks, e.g., cooperative perception, collaborative mapping, joint motion forecasting, and coordinated path planning. 
(3) It offers rich diversity, including different sensor types, driving environments, and ego vehicle trajectories.

%

%

\subsection{Limitations of Prior Works}

Recent advancements in autonomous driving have demonstrated the effectiveness of end-to-end planning frameworks, e.g., UniAD~\cite{hu2023planning}, VAD~\cite{jiang2023vad}, and Drive-VM~\cite{wang2024driving}.
Meanwhile, CAD has gained traction, with most efforts focusing on perception~\cite{hong2024multi,zhang2024ermvp}.
Comprehensive study of cooperative autonomy, however, is still limited by several key factors.
First, existing real-world datasets such as KITTI~\cite{kitti}, nuScenes~\cite{caesar2020nuscenes}, and DAIR-V2X~\cite{DAIR-V2X2021} are either designed for single-vehicle settings or constrained by limited sensor setups and small-scale collaborations. For instance, the V2V4Real~\cite{xu2023v2v4real} dataset involves only two vehicles, each with just two cameras and one LiDAR. Such limitations restrict their scalability and hinder comprehensive research on cooperative autonomous driving across multiple tasks.
Second, existing cooperative perception methods mostly focus on bird’s-eye-view (BEV) feature fusion~\cite{xu2022cobevt,liu2022bevfusion}, where dense feature maps are transmitted and aligned across vehicles. While effective, this strategy incurs high communication costs and potential redundancy, limiting the applicability under bandwidth constraints.
Third, existing autonomous driving datasets~\cite{xu2022opv2v, v2x-sim} created in simulation do not provide a clear pathway for transferring to real-world benchmarks. As a result, current cooperative methods validated solely in simulation cannot be reliably assessed for their effectiveness in real-world scenarios. This sim-to-real gap not only hinders fair evaluation but also limits the practical deployment of cooperative solutions.

\subsection{Contributions}

\textbf{M$^3$CAD.} To overcome these issues, we introduce a novel benchmark designed specifically to support research in \underline{\textbf{M}}ulti-vehicle, \underline{\textbf{M}}ulti-task, and \underline{\textbf{M}}ulti-modality \underline{\textbf{C}}ooperative \underline{\textbf{A}}utonomous \underline{\textbf{D}}riving (\textbf{M$^3$CAD}). 
By leveraging the advanced rendering capabilities of the recently-released Unreal Engine 5 (UE5) in CARLA~\cite{dosovitskiy2017carla}, we can simulate realistic multi-vehicle interactions within diverse driving scenarios.
%
M$^3$CAD comprises 204 sequences in total, offering over 30k frames and more than 267k annotated instances, along with the ground truth data for multiple autonomous driving tasks. 
%
Each sequence includes 10-60 collaborative vehicles with their precise location and trajectory information at each timestamp, as well as the map and occupancy details.
This comprehensive dataset supports a wide range of autonomous driving tasks, including object detection and tracking, mapping, motion forecasting, occupancy, and path planning, addressing the limitations in the single-vehicle end-to-end benchmarks (e.g., nuScenes~\cite{caesar2020nuscenes}) and cooperative non-end-to-end datasets (e.g., OPV2V~\cite{xu2022opv2v}). To the best of our knowledge, M$^3$CAD is currently the most comprehensive benchmark for both single-vehicle and cooperative autonomous driving research, while supporting more realistic vehicle movements and interactions in complex environments.

\begin{table*}[!t]
    \setlength{\tabcolsep}{3pt}
    \renewcommand{\arraystretch}{1.3}
    \caption{Detailed comparison of M$^3$CAD with existing benchmarks. Traffic: realistic and interactive traffic such as merging, lane crossing and traffic jam. Human-like: human like non-playable characters (NPCs) instead of rule based simulation. MP: Mapping, MF: Motion Forecasting, OCC: Occupancy Prediction, PP: Path Planning. Since perception tasks such as detection and tracking are standard components across datasets, they are not included in this comparison.}
    \label{tab:relatedwork}
    \centering
    \begin{tabular}{|l|c|c|ccc|cc|cccc|}
    \hline
    \multirow{3}{*}{\textbf{Benchmarks}} & \multirow{3}{*}{\textbf{3D Labels}} & \multirow{3}{*}{\textbf{Source}} &
    \multicolumn{3}{c|}{\textbf{Diverse Scenarios}} &
    \multicolumn{2}{c|}{\textbf{Diverse Cooperation}} &
    \multicolumn{4}{c|}{\textbf{Multiple Tasks}} \\
    & & & \textbf{Night / Rain} & \textbf{Human-like} & \textbf{Pedestrian} &
    \textbf{Coop. Vehicles} & \textbf{Coop. Range ($m$)} &
    \textbf{MP} & \textbf{MF} & \textbf{OCC} & \textbf{PP} \\
    \hline
    nuScenes~\cite{caesar2020nuscenes} & 1.4M & real & $\checkmark$/$\checkmark$ &$\checkmark$ & $\checkmark$ &
    N/A & N/A & $\checkmark$ & $\checkmark$ & $\checkmark$ & $\checkmark$ \\ 
    OPV2V~\cite{xu2022opv2v} & 232K & sim & $\times$/$\times$ & $\times$ & $\times$ &
    2--7 & 120 & $\checkmark$ & $\times$ & $\times$ & $\times$ \\ 
    V2X-Sim~\cite{v2x-sim} & 26.2K & sim & $\times$/$\times$ & $\times$ & $\times$ &
    2--5 & 70 & $\times$ & $\times$ & $\times$ & $\times$ \\ 
    V2V4Real~\cite{xu2023v2v4real} & 240K & real & $\times$/$\times$ & $\checkmark$ & $\checkmark$ &
    2 & 200 & $\times$ & $\checkmark$ & $\times$ & $\times$ \\ 
    V2X-Seq~\cite{v2x-seq} & 10.45K & real & $\checkmark$/$\times$  & $\checkmark$& $\checkmark$ &
    2--4 & 280 & $\checkmark$ & $\checkmark$ & $\times$ & $\times$ \\ 
    TUMTraf V2X~\cite{zimmer2024tumtraf} & 29.38K & real & $\checkmark$/$\times$ & $\checkmark$ & $\checkmark$ &
    2--4 & 200 & $\checkmark$ & $\checkmark$ & $\times$ & $\times$ \\ 
    V2X-Real~\cite{v2x-real} & 1.2M & real & $\times$/$\times$ & $\checkmark$ & $\checkmark$ &
    4 & $-$ & $\times$ & $\checkmark$ & $\times$ & $\times$ \\ 
    DAIR-V2X~\cite{DAIR-V2X2021} & 464K & real & $\checkmark$/$\checkmark$ & $\checkmark$ & $\checkmark$ &
    2 & 200 & $\times$ & $\times$ & $\times$ & $\times$ \\
    V2XSet~\cite{xu2022v2x}  & 230K & sim & $\times$/$\times$ & $\checkmark$ & $\checkmark$ &
    2-7 & 280 & $\times$ & $\times$ & $\times$ & $\times$ \\
    V2X-Radar~\cite{Huang2024V2X-R}  & 350K & real & $\checkmark$/$\checkmark$ & $\checkmark$ & $\checkmark$ &
    2 & 200 & $\times$ & $\times$ & $\times$ & $\times$ \\
    WHALES~\cite{wang2025whales}  & 2.01M & sim & $\times$/$\times$ & $\times$ & $\checkmark$ &
    8.4 & 200 & $\times$ & $\times$ & $\times$ & $\times$ \\
    \hline
    \textbf{M$^3$CAD (ours)} & \textbf{267K} & \textbf{sim} & $\checkmark$/$\checkmark$ & $\checkmark$ & $\checkmark$ &
    \textbf{10-60} & \textbf{200} & $\checkmark$ & $\checkmark$ & $\checkmark$ & $\checkmark$ \\
    \hline
    \end{tabular}
    \vspace{-3mm}
\end{table*}

\textbf{Multi-Level Fusion.} 
%
While M$^3$CAD provides the foundation for evaluating cooperative perception in complex scenarios, existing methods are mostly based on dense BEV feature fusion~\cite{xu2022cobevt,liu2022bevfusion}, resulting in high communication costs and poor scalability.
To close this gap, we propose a multi-level fusion method that adaptively balances communication efficiency and perception accuracy. Specifically, our framework explores three complementary strategies: \textit{BEV Feature Fusion}, which fuses dense feature maps to provide spatial information; \textit{Query Fusion}, which fuses compact, trajectory-aware features that preserve temporal and sequence information; and \textit{Reference Points Fusion}, which shares only sparse spatial to directly guide attention. By dynamically selecting the appropriate fusion level according to network conditions and system requirements, our framework enables scalable and efficient cooperative perception, maintaining strong accuracy while significantly reducing bandwidth consumption.

\textbf{Transfer to Real-World Benchmark.}
%
While multi-level fusion works well in simulations, it's unclear how it performs on real-world data. To explore this, we align M$^3$CAD with the nuScenes dataset and the cross-domain performance. Our results show that UniAD~\cite{hu2023planning}, when pre-trained on M$^3$CAD, can be effectively fine-tuned with only 10\% of nuScenes data, leading to significant performance improvements in real-world settings.

\textbf{Extensive Evaluations}. Finally, we conduct comprehensive experiments to validate our framework. These include: (1) systematic benchmarking across multiple cooperative driving tasks, (2) analysis of communication bandwidth requirements to assess the efficiency of the proposed multi-level fusion, and (3) robustness tests under ego-pose and sensor noises. These evaluations demonstrate the effectiveness, robustness, and generalizability of our M$^3$CAD benchmark.

\section{Related Works}

\textbf{Benchmarks for A multi-Task Autonomous Driving.}
Multi-task autonomous driving framework jointly optimizes multiple modules for individual tasks while prioritizing the ultimate planning objective, benefiting from enhanced safety and interpretability~\cite{chen2024end}. 
To support multi-task research, the nuScenes benchmark~\cite{caesar2020nuscenes} is proposed and is recognized as the most valuable resource in autonomous driving research, due to its comprehensive and high-quality data that encompasses a wide range of multi-task driving scenarios. 
Several end-to-end autonomous driving solutions have been evaluated using the nuScenes dataset, e.g., DiffStack~\cite{karkus2023diffstack}, MP3~\cite{casas2021mp3}, P3~\cite{sadat2020perceive}, ST-P3~\cite{hu2022st}, and VAD~\cite{jiang2023vad}. 
Among them, a notable example is UniAD~\cite{hu2023planning}, a unified framework that seamlessly integrates perception, prediction, and planning into a single architecture. 
While multi-task end-to-end autonomous driving frameworks have shown promising performance by jointly optimizing all modules, there remains a significant gap in exploring collaborative strategies among multiple vehicles for autonomous driving tasks.

\textbf{Benchmarks for Cooperative Perception.}
Despite the progress in end-to-end autonomous driving, multi-vehicle collaboration has primarily focused on cooperative perception, leading this field underexplored. 
To support this direction, several benchmarks have been introduced, including OPV2V~\cite{xu2022opv2v}, V2V4Real~\cite{xu2023v2v4real}, V2X-Seq~\cite{v2x-seq}, TUMTraf-V2X~\cite{zimmer2024tumtraf}, DAIR-V2X-C~\cite{DAIR-V2X2021}, V2X-Set~\cite{xu2022v2x}, V2X-Radar~\cite{Huang2024V2X-R}, and WHALES~\cite{wang2025whales}. 
However, these datasets are largely centered around perception tasks (particularly object detection), which limits their utility for more comprehensive autonomous driving challenges such as object tracking, motion forecasting, and ultimately, path planning. 
Table~\ref{tab:relatedwork} presents a comparative summary of existing benchmarks and our proposed M$^3$CAD, highlighting the differences in scale of dataset, diversity of scenarios, cooperation settings, and types of supported autonomous driving tasks.
%

\section{The M$^3$CAD Benchmark}
\label{sec:ground truth}
M$^3$CAD is designed to serve as a versatile platform to facilitate various cooperative driving tasks. 
To achieve this goal, we deployed multiple vehicles in the Town01, Town02, Town03, Town04, and Town10HD maps in CARLA.
Each vehicle is equipped with six $1920\times1080$-pixel (110\textdegree{} FOV) cameras, a 64-beam LiDAR, and GPS/IMU sensors to collect multi-modality sensor data during daytime, nighttime, and various weather conditions.

\begin{figure*}[h]
  \centering
  \begin{subfigure}[b]{0.348\textwidth}
    \includegraphics[width=\textwidth]{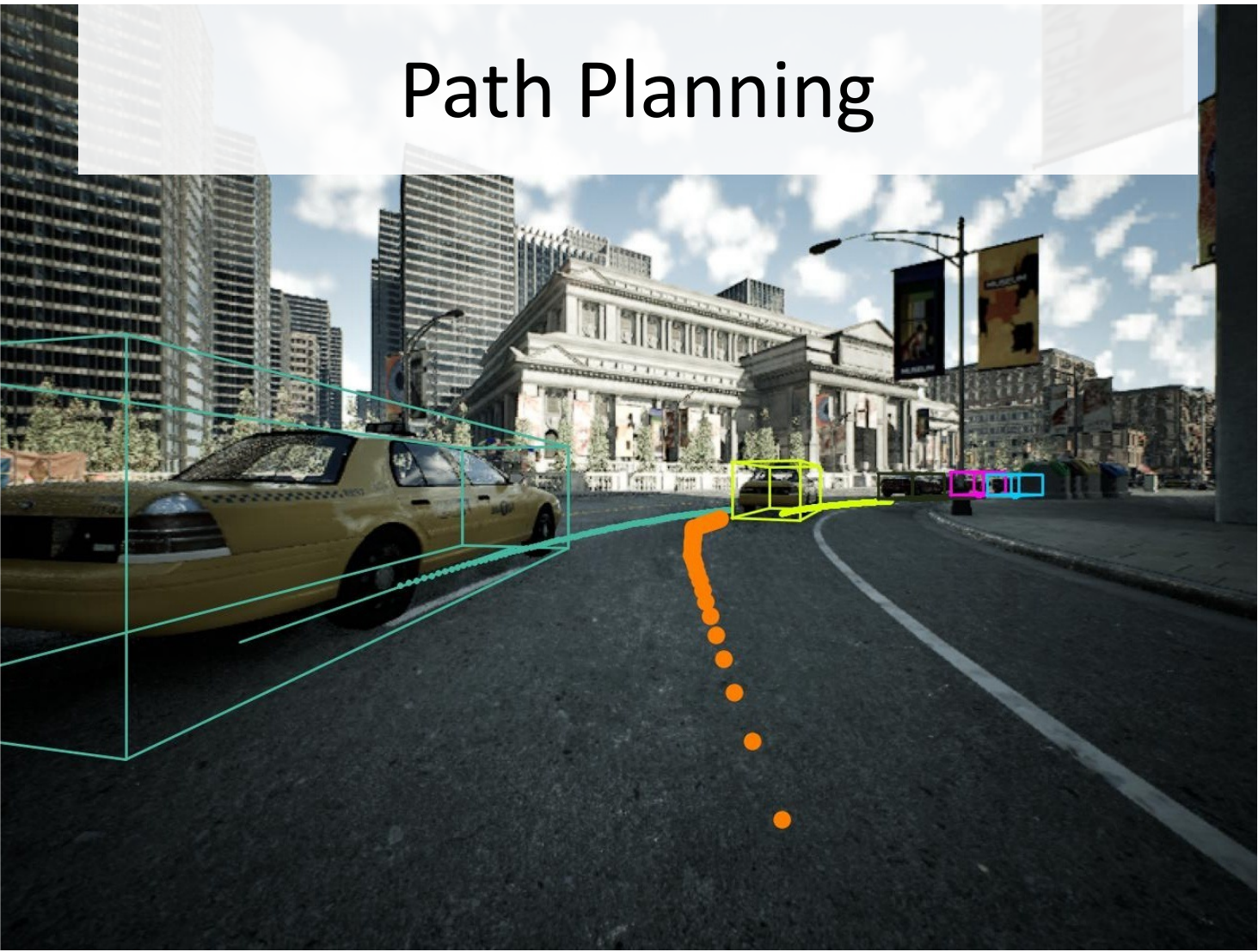}
    \caption{CAM\_FRONT}
    \label{fig:pp}
  \end{subfigure}
  \hfill
  \begin{subfigure}[b]{0.348\textwidth}
    \includegraphics[width=\textwidth]{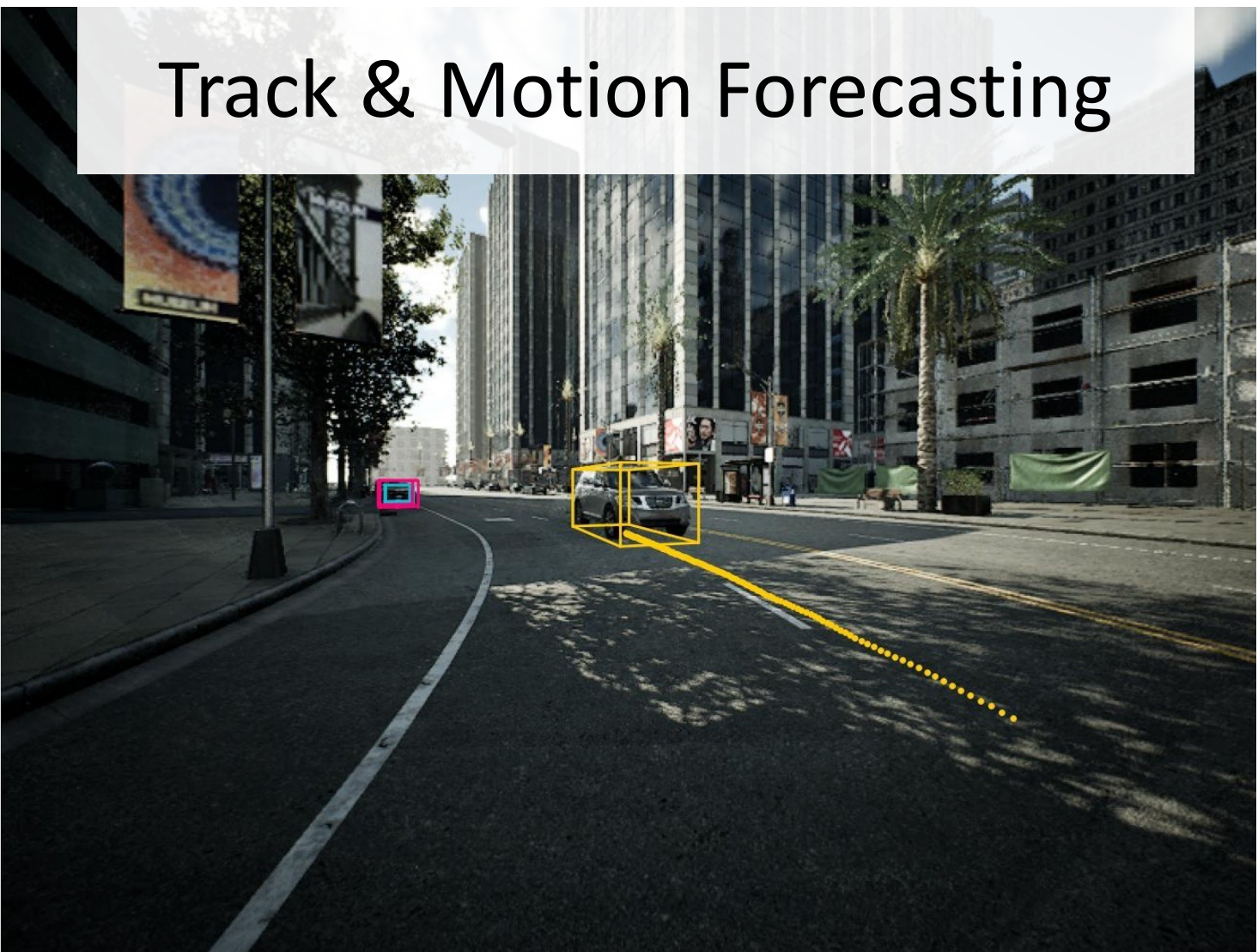}
    \caption{CAM\_BACK}
    \label{fig:ot-mf}
  \end{subfigure}
  \hfill
  \begin{subfigure}[b]{0.264\textwidth}
    \includegraphics[width=\textwidth]{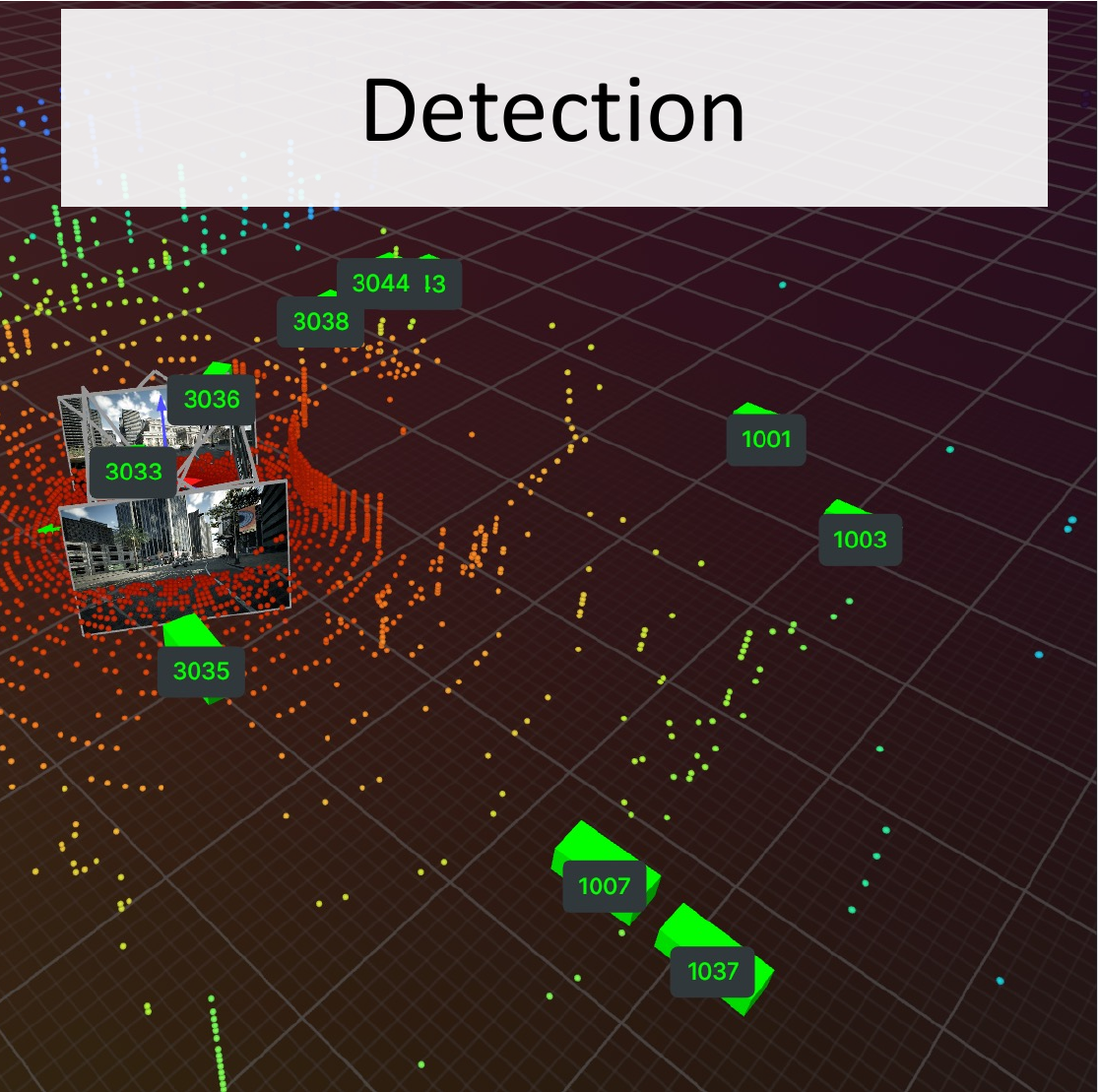}
    \caption{3D View}
    \label{fig:od}
  \end{subfigure}

  \vspace{0em}

  \begin{subfigure}[b]{0.348\textwidth}
    \includegraphics[width=\textwidth]{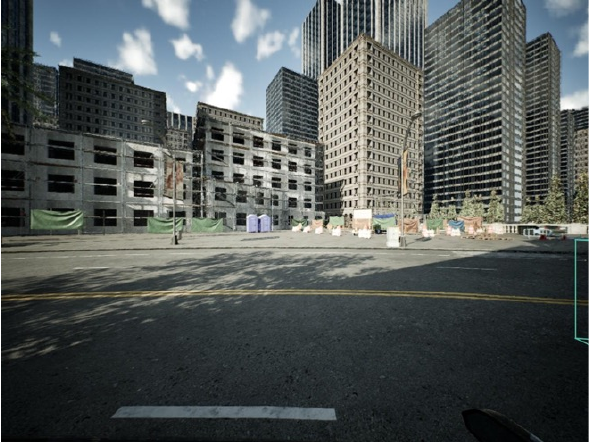}
    \caption{CAM\_LEFT}
    \label{fig:camer_left}
  \end{subfigure}
  \hfill
  \begin{subfigure}[b]{0.348\textwidth}
    \includegraphics[width=\textwidth]{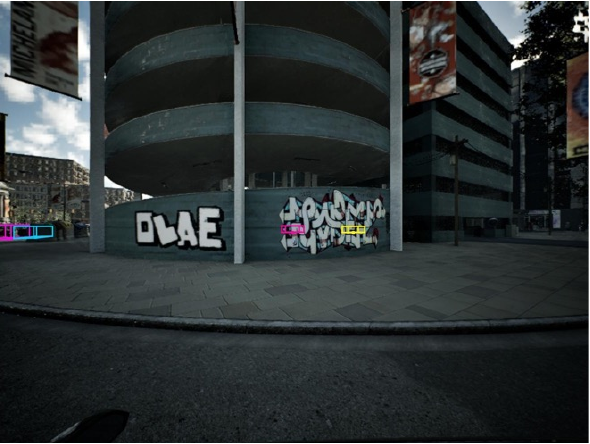}
    \caption{CAM\_RIGHT}
    \label{fig:cam_right}
  \end{subfigure}
  \hfill
  \begin{subfigure}[b]{0.264\textwidth}
    \includegraphics[width=\textwidth]{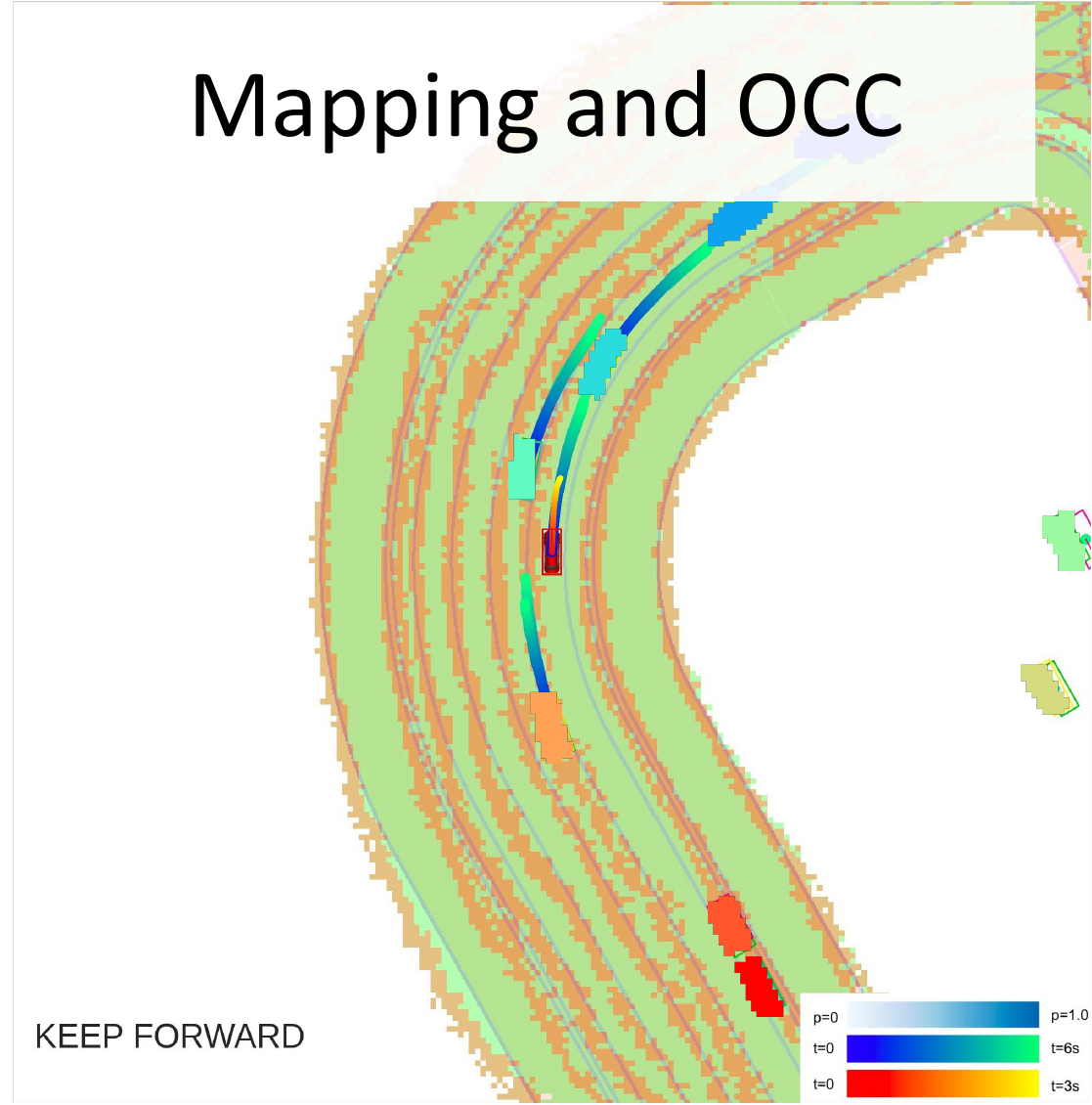}
    \caption{BEV}
    \label{fig:map&occ}
  \end{subfigure}
  \caption{Illustrations of various autonomous driving tasks using the M$^3$CAD dataset. (a) Demonstrates the path planning ({PP}) results, where the ego vehicle's predicted trajectory is represented by a dotted line. (b) Shows object tracking ({OT}) and motion forecasting ({MF}) results where dotted lines represent predicted trajectories of other vehicles. (c) Presents object detection  ({OD}) results in 3D space. (f) Depicts mapping  ({MP}) and occupancy prediction ({OCC}) results.}
  \vspace{-5mm}
  \label{fig:examples}
\end{figure*}

\textbf{Annotation.}
We annotate the collected data based on the nuScenes format~\cite{caesar2020nuscenes}, enabling research on multi-vehicle, multi-task and multi-modality autonomous driving. 
%
To support research on cooperative driving tasks, we provide in M$^3$CAD  comprehensive annotations for {object detection}, {tracking}, {motion prediction}, {occupancy forecasting}, and {path planning}. 
All objects located within the $102.4 m \times 102.4 m$ BEV range centered on the ego vehicle are annotated, including attributes: {location}, {speed}, {bounding box center}, and {extent} ({length}, {width}, {height}).
To support effective collaboration between vehicles, M$^3$CAD also provides transformation matrices that align the ego vehicle with any collaborating vehicles, enabling accurate information fusion.
For object detection and tracking tasks, only objects within the BEV range are considered to reflect realistic constraints. The dataset includes motion histories and future trajectories for vehicles, as well as binary BEV occupancy maps generated from LiDAR point clouds to support occupancy forecasting. Vehicle trajectories are also recorded to assist the planning task. Additionally, the semantic map contains multiple layers, e.g., drivable areas, lane dividers, and road dividers, all aligned with the CARLA global coordinate system and structured according to the nuScenes specification. 

\textbf{Dataset Split and Multiple Tasks.} The M$^3$CAD dataset is divided into training, validation, and test subsets using a 70/15/15 split. The rich data generated by these vehicles allows for the exploration of various tasks.
These tasks include object detection, object tracking, mapping, motion forecasting, occupancy prediction, and, critically, path planning.
Fig.~\ref{fig:examples} shows the results of UniAD's different tasks performed on the M$^3$CAD dataset.

\begin{figure}[H]
    \centering
    \vspace{-2.5mm}
    \begin{minipage}{0.48\linewidth}
        \centering
        \includegraphics[width=\linewidth]{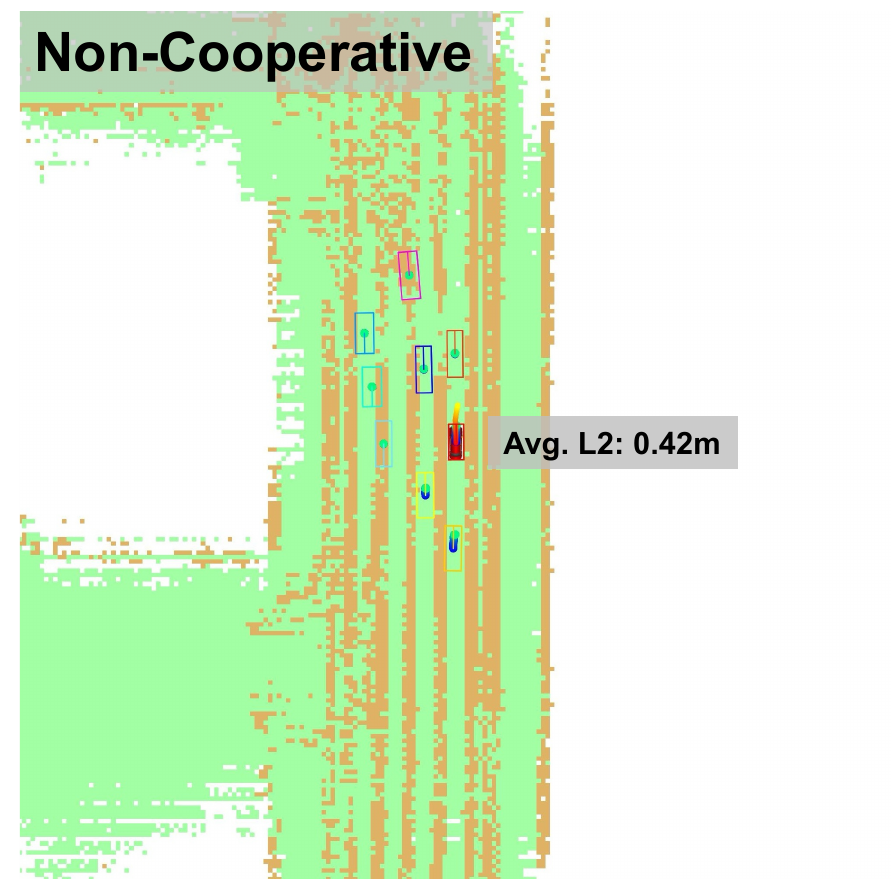}
        \caption*{(a) Non-Cooperative}
    \end{minipage}
    \hfill
    \begin{minipage}{0.48\linewidth}
        \centering
        \includegraphics[width=\linewidth]{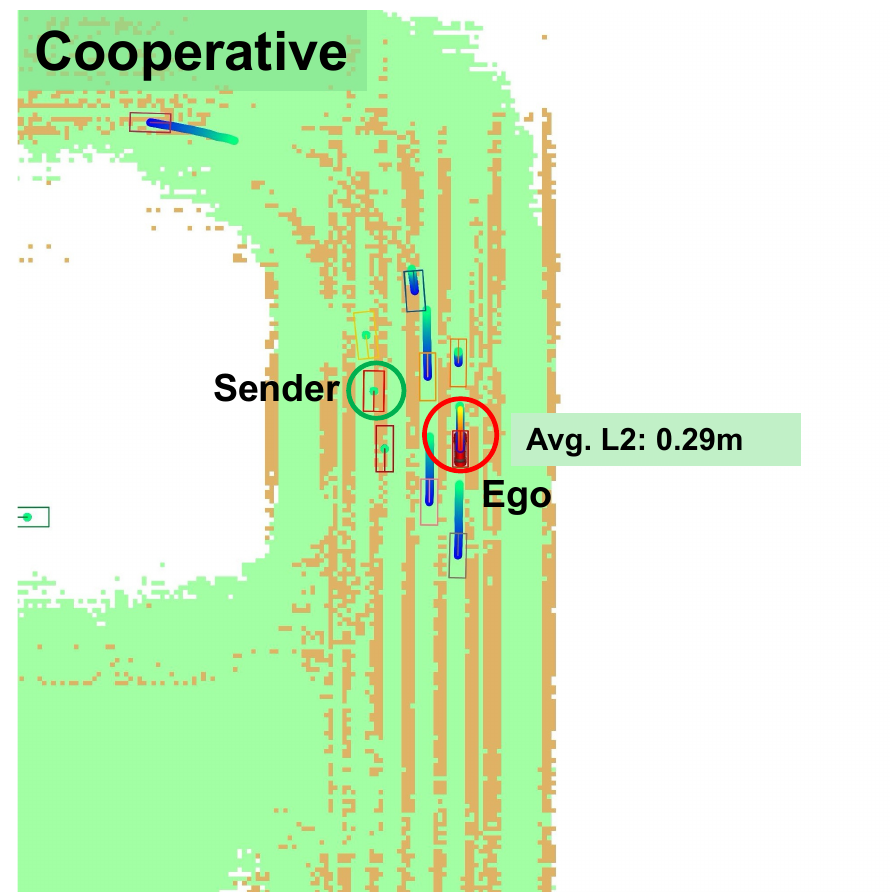}
        \caption*{(b) Cooperative}
    \end{minipage}
    \caption{
    Qualitative comparison between cooperative and non-cooperative path planning. 
    All visualizations are shown in the BEV, with each vehicle depicted as a uniquely colored box.
    The blue-green curves represent the predicted trajectories of vehicles over the next 6 seconds, showing only the top-1 predicted trajectories here.
    The red-yellow curves present the planned paths of the ego vehicle within a 3-second window.
    In the underlying semantic map, green areas indicate drivable regions and yellow dots mark lane dividers.
    }
    \label{fig:cooper}
\end{figure}

One of the prominent features of M$^3$CAD is its support for multi-vehicle collaboration, enabling cooperative path planning to enhance driving safety.
Although cooperative object detection has been extensively studied~\cite{fcooper,xu2022opv2v, xu2022cobevt, vnet, qu2024sicp, qu2024head}, other tasks such as cooperative mapping, motion forecasting, occupancy prediction, and planning remain much less explored. Recent studies have explored cooperative motion planning~\cite{zhang2025comtp, wang2025cmp}; however, a comprehensive understanding of how different tasks interact and depend on each other is still missing.
As shown in Fig.~\ref{fig:cooper}, with the assistance from the sender vehicle through a BEV fusion method, the ego vehicle can obtain a better perception of the surrounding environment, allowing it to follow a trajectory much closer to the ground truth (with a L2 error of $0.29m$).
Without M$^3$CAD, it would be extremely difficult, if not impossible, to systematically evaluate the benefits and trade-offs of different tasks within cooperative autonomous driving settings.
%

\vspace{-5mm}
\section{Multi-level Fusion}
\begin{figure}[H]
    \centering
\includegraphics[width=0.62\textwidth]{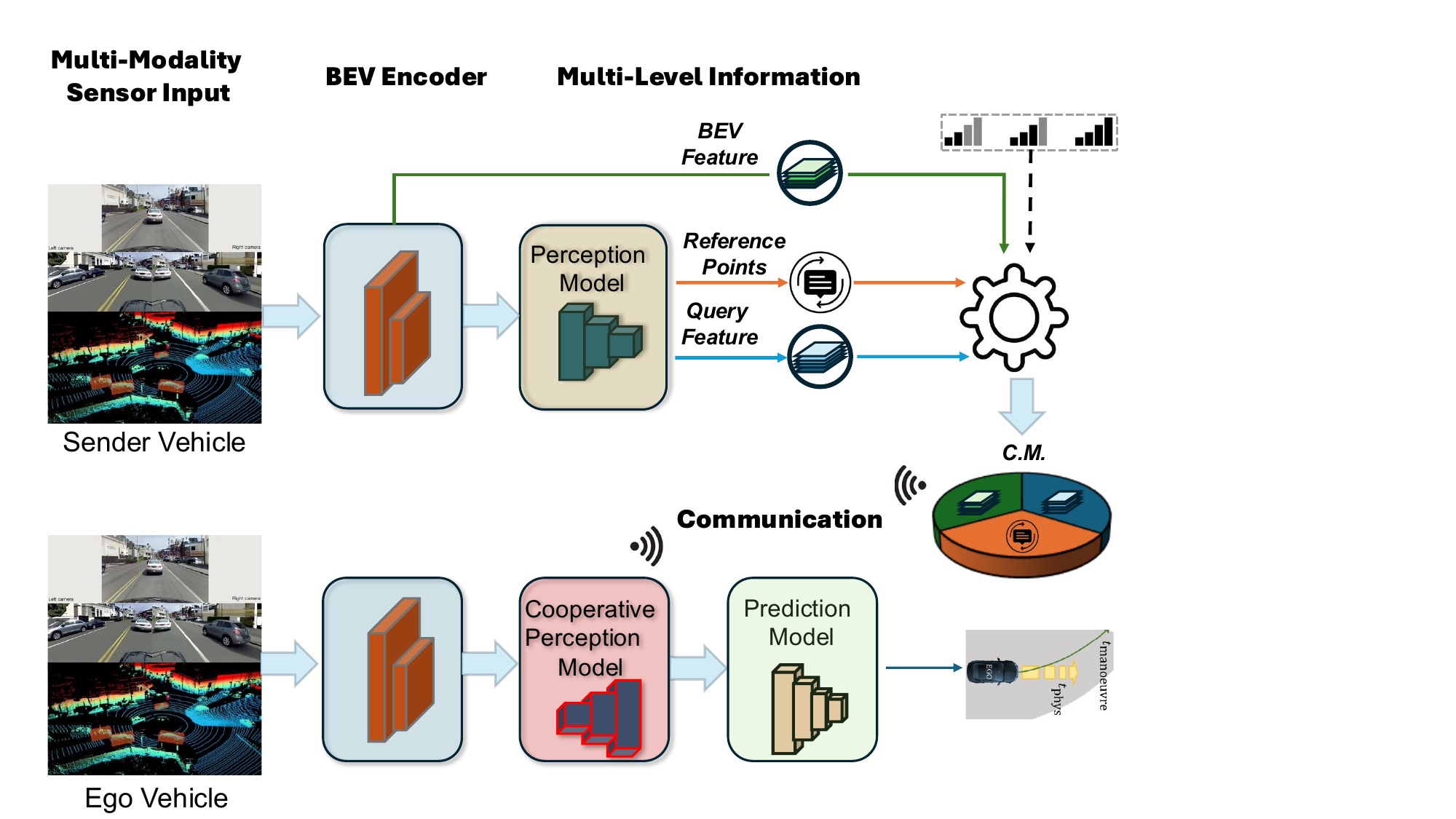}
    \caption{Multi-level cooperative perception on CAVs. Sender vehicles generate BEV features, queries, and reference points, which are selectively packaged into a cooperative message (C.M.) based on bandwidth and task requirements. The ego vehicle receives these messages and performs multi-level fusion to achieve cooperative perception.}
    \label{fig:arch}
\end{figure}

To better understand how cooperative perception improves the ego vehicle's performance, we now take a closer look at what information can be shared between CAVs and how this sharing happens.
Unlike existing approaches, which are commonly divided into high-level, intermediate-level~\cite{fcooper,wang2024emiff,huang2024actformer}, and low-level sharing~\cite{chen2019cooper}, we propose a unified multi-level information sharing framework to support both high- and intermediate-level cooperative perception.
%
As shown in Fig.~\ref{fig:arch}, the input to our framework can be multi-modal sensor data. In this work, we instantiate it with camera inputs, which will be processed by various perception modules e.g., BEVFormer~\cite{li2022bevformer} or MOTR~\cite{zeng2022motr}, to obtain different types of data: BEV features, query features, and reference points.
These outputs are then passed to the prediction model to generate the final path planning results.

%
Relying on the BEVFormer, BEV features are extracted to represent the spatial and semantic layout of the environment from a bird’s-eye view. 
These features can be shared among vehicles to realize cooperative perception~\cite{xu2022cobevt,liu2022bevfusion}. 
Although effective, it suffers from high communication cost, limiting its real-world applications. 
To overcome this issue, we explore another way to represent the information, known as query features~\cite{univ2x,coopdetr}.
Modern transformer-based perception modules, e.g., UniAD-TrackFormer~\cite{hu2023planning}, can generate detailed track-level outputs, including queries, reference points, object indices, confidence scores, and predicted boxes. 
Among these, queries are the most informative while remaining compact.
They include not just spatial information, but also capture important details about time, motion, and object identity, making them especially useful for tracking tasks.
Queries from different vehicles can be fused to realize cooperative perception, which uses much less bandwidth than fusing BEV features.
To further reduce bandwidth requirements, we suggest that vehicles share only reference point information to achieve cooperative perception.



\subsection{BEV Feature Fusion (BFF)} 
In this paradigm, each vehicle transmits its own BEV feature maps to the ego vehicle for alignment and fusion. 
Assuming vehicle $v_j$ is selected for cooperation, it needs to send its BEV feature $\mathcal{F}{j}$ to the ego vehicle, along with its pose $\mathcal{P}{j}$ and location $\mathcal{L}{j}$ information. 
Based on the affine transformation function $T_j=\Psi(\mathcal{P}{j}$, $\mathcal{L}{j})$, the  transformation matrix $T_j$ can be obtained.
Then, the ego vehicle transforms the received feature map to align with its perspective, resulting in a transformed feature map $\mathcal{F}{j}^{'}$.
The transformed $\mathcal{F}{j}^{'}$ is then fused into the ego's as follows: 
\begin{equation*} \mathcal{F}^* = \Phi(\mathcal{F}{e} \parallel \mathcal{F}{j}^{'}) \in \mathbb{R}^{H \times W} \tag{1} \end{equation*} 
where $\mathcal{F}{e}$ is the ego vehicle's BEV feature, $H$ and $W$ are the height and width of the feature map, and $\Phi(\cdot)$ denotes the fusion function.
A variety of methods can implement the function $\Phi$, e.g., those in F-Cooper~\cite{fcooper}, Attentive Fusion~\cite{xu2022opv2v}, CoBEVT~\cite{xu2022cobevt}, V2VNet~\cite{vnet}, Where2Comm~\cite{hu2022where2comm}, V2VAM~\cite{li2023learning}, V2X-ViT~\cite{xu2022v2x}, CoAlign~\cite{lu2023robust}, and SiCP~\cite{qu2024sicp}. 
Within the M$^3$CAD benchmarks, several above-mentioned fusion strategies are provided, offering the flexibility of selecting or extending them with new fusion methods.


%

\subsection{Query Feature Fusion (QFF)}
While BFF is effective, it requires sending a lot of data. 
To reduce this cost, we propose query feature fusion (QFF), which uses smaller query features to achieve cooperative perception.
Different perception modules produce queries with different meanings, e.g., object queries generated by DETR~\cite{detr} represent potential objects in the scene, while track queries produce by UniAD-TrackFormer represent tracked objects across time.
Here, we use track query as an example to illustrate how cooperative perception can be achieved on the tracking task.


\begin{table*}[h]
    \caption{Comparison of multi-level cooperative methods performance on various autonomous driving tasks.}
     \centering
     \resizebox{0.98\textwidth}{!}{
    \begin{tabular}{c|ccc|cc|cc|cc|c}
    \hline
    \multicolumn{1}{c|}{\textbf{}} & \multicolumn{3}{c|}{\textbf{Tracking}} & \multicolumn{2}{c|}{\textbf{Mapping ($\%$)}} & \multicolumn{2}{c|}{\textbf{Motion Forecasting ($m$)}} & \multicolumn{2}{c|}{\textbf{Occupancy Prediction ($\%$)}} & \multicolumn{1}{c}{\textbf{Planning ($m$)}}\\
    \textbf{Method} & {AMOTA$\uparrow$} & {AMOTP$\downarrow$} & {Recall$\uparrow$} & {IoU-Lane$\uparrow$} & {IoU-Road$\uparrow$} & {ADE$\downarrow$} & {FDE$\downarrow$} & {IoU-n$\uparrow$} & {IoU-f$\uparrow$} & {L2$\downarrow$}\\ 
    \hline
    No fusion & 0.21 & 0.66 & 0.48 & 49.6 & 94.0 & 0.36 & 0.38 & 76.2 & 57.5 & 0.43 \\
    F-cooper & 0.720 & 0.680 & 0.816 & - & - & - & - & - & - & - \\
    \textbf{RPF} & 0.671 & 0.684 & 0.758 & \textbf{58.3} & \textbf{96.0} & 0.346 & 0.358 & 79.5 & 62.5 & 0.300 \\
    \textbf{QFF} & 0.697 & 0.601 & 0.835 & 55.6 & 95.6 & 0.3490 & 0.3627 & 80.5 & \textbf{63.8} & \textbf{0.221} \\
    \textbf{BFF} & \textbf{0.774} & \textbf{0.579} & \textbf{0.846} & 50.67 & 95.34 & \textbf{0.2797} & \textbf{0.2976} & \textbf{80.9} & 63.0 & 0.3109 \\
    \hline
    \end{tabular}}
    \label{tab:multi-level methods comparision}
\end{table*}

In UniAD-TrackFormer, each tracked object is represented by a track query, which includes a learnable embedding, reference points, a class score, and a bounding box.
These queries are passed from frame to frame using query interactions, helping the model keep track of the same object over time.
%
To enable track query fusion, we first align the sender's track instances into the ego's coordinate system, using extrinsic calibration information. 
Next, we combine the matching query embeddings using a Multi-Layer Perceptron (MLP):
\begin{equation}
    \tilde{q}_i = \text{MLP}\big([q_{i,\text{ego}} \,\|\, q_{j,\text{sender}\rightarrow ego}]\big),
\end{equation}
where $[ \cdot \| \cdot ]$ denotes concatenation. 
Finally, the fused queries $\tilde{q}_i$ are passed into the decoder on the ego vehicle to update its tracking results.

%
Note that the QFF mechanism can also be used for other perception tasks by fusing queries generated from their corresponding transformer-based modules.
QFF greatly reduces communication compared to BFF; however, query embeddings are still high-dimensional, making them costly to transmit over current networks.
To further cut down bandwidth requirements, while keeping important spatial information, we introduce reference points fusion.

\subsection{Reference Point Fusion (RPF)} 
The key idea of RPF is is to move from sharing large amounts of dense feature data to sharing smaller, more meaningful, high-level information.
The reference points from UniAD-TrackFormer show where potential tracked objects might be.
By using reference points shared from other vehicles, the ego vehicle can improve its own tracking performance.
Specifically, the ego vehicle combines its own reference points with those received from senders. 
Formally, let $\mathcal{R}_{ego}^t$ and $\mathcal{R}_{sender \rightarrow ego}^t$ 
denote the ego’s and sender’s reference point sets at frame $t$. 
The fused set is obtained as
\[
    \tilde{\mathcal{R}}^t = \mathcal{R}_{ego}^t \cup 
    \Big( \mathcal{R}_{sender \rightarrow ego}^t \setminus \mathcal{R}_{ego}^t \Big).
\]

This ensures ego preserves all self-derived reference points while enlarging its search space (within BEV) with more priors contributed by the sender. 
By transmitting only sparse information reference point fusion thus yields the minimal communication cost, making it highly practical for current vehicular networks.
By combining BFF, QFF, and RPF, we build a progressive multi-level framework that adaptively balances perception accuracy and communication cost.

\section{Experiments}
To test how well our multi-level fusion framework works, we evaluate it on multiple cooperative perception tasks using the M$^3$CAD dataset.
\subsection{Multiple Tasks for Cooperative Perception and Prediction}
%
We compare the performance of several perception and prediction tasks using no fusion, early fusion \cite{fcooper}, and our proposed multi-level fusion. 
The tasks include tracking, mapping, motion forecasting, occupancy prediction, and planning.
The detailed results for all methods are shown in Table~\ref{tab:multi-level methods comparision}, showcasing the different performances of our multi-level fusion strategies.
Since BFF sends the most data, it achieves the best or second-best results in almost every task, which explains why most existing cooperative perception methods are BFF-based. 
Compared to no fusion, however, BFF's improvement is significant, showing how important cooperative perception really is.
Interestingly, we found that RPF also performs very well, i.e., its L2 planning error is only $8 cm$ larger ($0.300 m$ vs. $0.221 m$) than QFF, which is not a too big difference. 
Notably, RPF performs best in mapping tasks, likely because reference points provide precise spatial information, which is especially important for tasks that need high accuracy like mapping.

\begin{figure}
    \centering
    \includegraphics[width=1\linewidth]{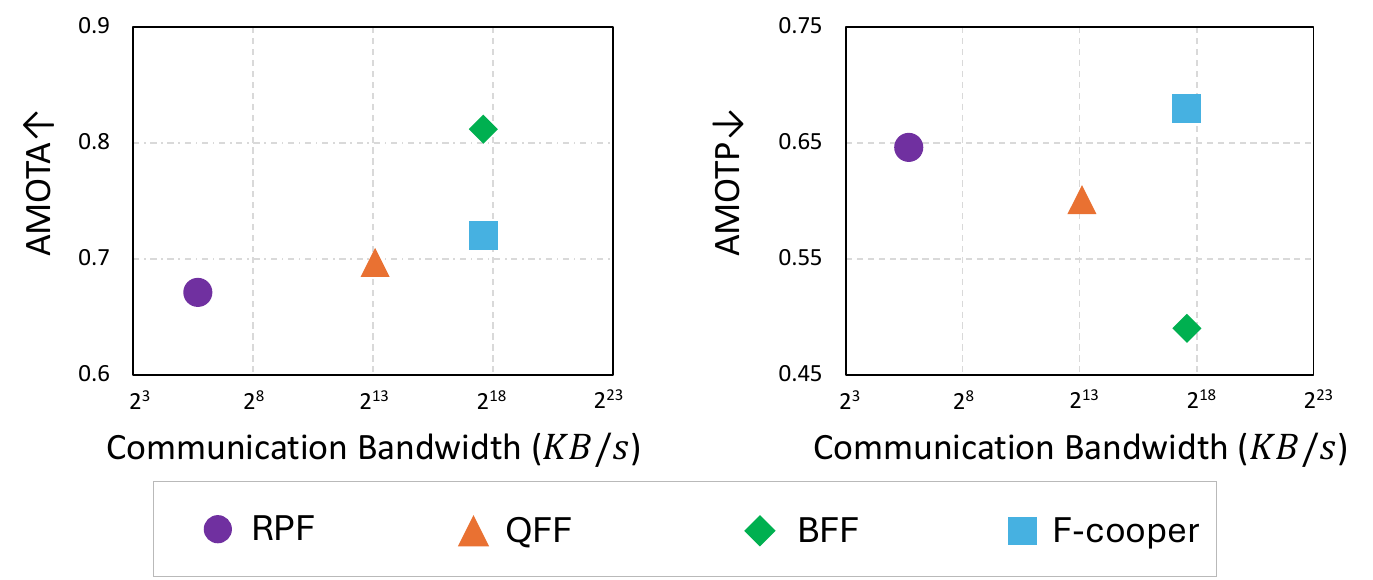}
    \caption{Comparison of cooperative tracking performance of different multi-level fusion methods and the F-Cooper method. It shows how much communication bandwidth each method needs, using a $log_2$ scale. QFF requires approximately 9,063 KB/s, BFF requires 200,000 KB/s, while RPF requires only 53 KB/s. These values are based on \textit{float32} precision and a 5 FPS transmission rate.}
    \label{fig:communication_cost}
\end{figure}
\subsection{Network Bandwidth Requirements} 
While the above results demonstrate the benefits of cooperative perception across different tasks, real-world deployment also needs to take communication costs into account.
To explore this, we compare how well different information fusion strategies perform in cooperative tracking, based on how much network bandwidth they require.
As illustrated in the Fig.~\ref{fig:communication_cost}, each fusion method has its own strengths.
RPF works best in low-bandwidth situations where it transmits minimal amount of data, about $1/3800$ of what BFF needs.
BFF gives the best tracking accuracy but uses the most bandwidth, making it ideal when performance matters more than bandwidth usage.
QFF strikes a balance, offering strong tracking performance while keeping communication costs reasonable.

\subsection{Benchmarking on Real-World Dataset}

To demonstrate that models trained on  M$^3$CAD can transfer effectively to real-world datasets, we evaluate the well-known multi-task autonomous driving model UniAD on the nuScenes benchmark after pretraining it on  M$^3$CAD. 
We compare three training strategies to test the sim-to-real transfer: (1) UniAD trained directly on $100\%$ of nuScenes data, (2) BEVFormer pretrained and then fine-tuned on $10\%$ of nuScenes data, and (3) UniAD pretrained on $100\%$ of  M$^3$CAD and then fine-tuned on $10\%$ of nuScenes data.

As shown in Table~\ref{tab:sim2real}, the M$^3$CAD pretraining approach achieves significant improvements over the baseline BEVFormer approach. 
Specifically, pretraining on M$^3$CAD reduces the average trajectory error from $1.91 m$ to $1.30 m$ (a $32\%$ improvement) and lowers the collision rate from $1.3\%$ to $0.57\%$ (a $56\%$ reduction) when fine-tuned with only $10\%$ nuScenes data. 
These results demonstrate that our synthetic dataset provides effective pretraining that substantially improves performance in data-limited real-world scenarios, validating the quality and transferability of M$^3$CAD.

\begin{table}[h]
\centering
\caption{We evaluate planning performance of UniAD~\cite{hu2023planning} on the nuScenes dataset using different training datasets.}
\label{tab:sim2real}
 \resizebox{\linewidth}{!}{
\begin{tabular}{lcc}
\toprule
\textbf{Training datasets} & \textbf{L2 $\downarrow$ ($m$)} & \textbf{Collision $\downarrow$ ($\%$)} \\
\midrule
100\% nuScenes   & 1.03 & 0.31 \\
10\% nuScenes & 1.91 & 1.3 \\
100\% M$^3$CAD + 10\% nuScenes  & 1.30 & 0.57 \\
\bottomrule
\end{tabular}}
\end{table}

\subsection{Impact of Noise in M$^3$CAD.} 

To make our M$^3$CAD dataset more realistic, we introduce two types of noise commonly seen in real-world autonomous driving: localization errors and calibration errors.
Specifically, we model the ego vehicle's localization errors follow a zero-mean Gaussian distribution. 
For the translational component, we apply a noise with standard deviations of $\sigma_x = 0.1 m$, $\sigma_y = 0.08 m$, and $\sigma_z = 0.02 m$, which reflects the typical accuracy characteristics of automotive-grade localization systems. 
For the rotational component, we introduce angular noise with standard deviations of $\sigma_{roll} = 0.2^{\circ}$, $\sigma_{yaw} = 1.0^{\circ}$, and $\sigma_{pitch} = 0.2^{\circ}$.
Similarly, we introduce camera calibration errors by adding zero-mean Gaussian noise to the sensor-to-LiDAR transformation. 
The rotational errors have standard deviations of $\sigma_{roll} = 0.1^{\circ}$, $\sigma_{pitch} = 0.1^{\circ}$, and $\sigma_{yaw} = 0.2^{\circ}$, while the translational errors have standard deviations of $\sigma_x = 0.01 m$, $\sigma_y = 0.01 m$, and $\sigma_z = 0.02 m$. 
For the camera intrinsics, we perturb the focal lengths ($f_x$, $f_y$) by $\pm 2.0$ pixels and the principal point coordinates ($c_x$, $c_y$) by $\pm 1.0$ pixel.


As shown in Table~\ref{tab:cooperative_noise_analysis}, under only localization noise, our cooperative perception solution remains robust. 
The mAP decreases from $0.785$ to $0.666$ (a $15\%$ drop), which is acceptable given typical GPS accuracy limits. 
AMOTA falls from $0.774$ to $0.664$ (a $14\%$ drop), showing some impact on tracking but still maintaining reliable performance for safe operation. 
Most importantly, planning remains highly resilient, with the L2 error rising only from $0.31 m$ to $0.49 m$, confirming that the system can still ensure safe navigation under realistic localization uncertainties.


Sensor calibration drift shows similar impact patterns, as shown in table~\ref{tab:cooperative_noise_analysis}, with mAP decreasing to $0.643$ and AMOTA to $0.628$, representing an $18\%$ and $19\%$ degradation, respectively. 
Interestingly, calibration errors have a more pronounced effect on perception tasks than localization errors, as evidenced by the slightly larger performance drops. 
However, the planning module demonstrates consistent robustness, with L2 error increasing to only $0.39 m$, confirming that our multi-level fusion strategy provides sufficient redundancy to maintain safe autonomous driving capabilities, despite sensor calibration uncertainties.

When both localization and calibration errors exist, our system experiences the expected cumulative effect with mAP dropping to $0.609$ and planning L2 error reaching $0.49 m$.
These results show that although individual error sources cause some performance drop, our cooperative framework still performs at an acceptable level. 
The relatively small degradation under challenging conditions confirms the effectiveness of our approach for practical use in a real-world setting.

\begin{table*}[]
\caption{Comparison of cooperative perception and planning performance on M$^3$CAD with different types of noise. Type 1 noise: ego localization errors, Type 2 noise: sensor calibration drifts.}
\centering
\resizebox{\textwidth}{!}{
\begin{tabular}{c|cccc|cc|ccc|cccc|cc}
\hline
\multicolumn{1}{c|}{\textbf{}} & \multicolumn{4}{c|}{\textbf{Object Detection and Tracking}} & \multicolumn{2}{c|}{\textbf{Mapping (\%)}} & \multicolumn{3}{c|}{\textbf{Motion Forecasting ($m$)}} & \multicolumn{4}{c|}{\textbf{Occupancy Prediction ($\%$)}} & \multicolumn{2}{c}{\textbf{Plan ($m, \%$)}}\\
\textbf{Noise Type} & {mAP$\uparrow$} & {AMOTA$\uparrow$} & {AMOTP$\downarrow$} & {Recall$\uparrow$} & {IoU-Lane$\uparrow$} & {IoU-Road$\uparrow$} & {ADE$\downarrow$} & {FDE$\downarrow$} & {MR$\downarrow$} & {IoU-n$\uparrow$} & {IoU-f$\uparrow$} & {VPQ-n$\uparrow$} & {VPQ-f$\uparrow$} & { L2$\downarrow$} & {Col.$\downarrow$}\\ 
\hline
Baseline (No noise) & 0.785 & 0.774 & 0.579 & 0.846 & 50.67 & 95.34 & 0.2797 & 0.2976 & 0.0001 & 80.9 & 63.0 & 76.9 & 61.4 & 0.3109 & 0.02 \\
Type 1 noise & 0.666 & 0.664 & 0.803 & 0.749 & 47.46 & 94.24 & 0.3754 & 0.4097 & 0.0004 & 76.2 & 54.4 & 71.9 & 51.6 & 0.4906 & 0.13 \\
Type 2 noise & 0.643 & 0.628 & 0.846 & 0.709 & 48.25 & 94.48 & 0.3661 & 0.3982 & 0.0007 & 75.4 & 52.0 & 70.5 & 48.2 & 0.3910 & 0.08 \\
Type 1+2 noise & 0.609 & 0.623 & 0.888 & 0.710 & 47.41 & 94.22 & 0.4270 & 0.4579 & 0.0005 & 75.0 & 50.4 & 70.2 & 45.7 & 0.4914 & 0.0014 \\
\hline
\end{tabular}}
\label{tab:cooperative_noise_analysis}
\end{table*}

\begin{figure}[h]
    \centering
    \begin{subfigure}[b]{0.45\linewidth}
        \centering
        \includegraphics[width=\linewidth]{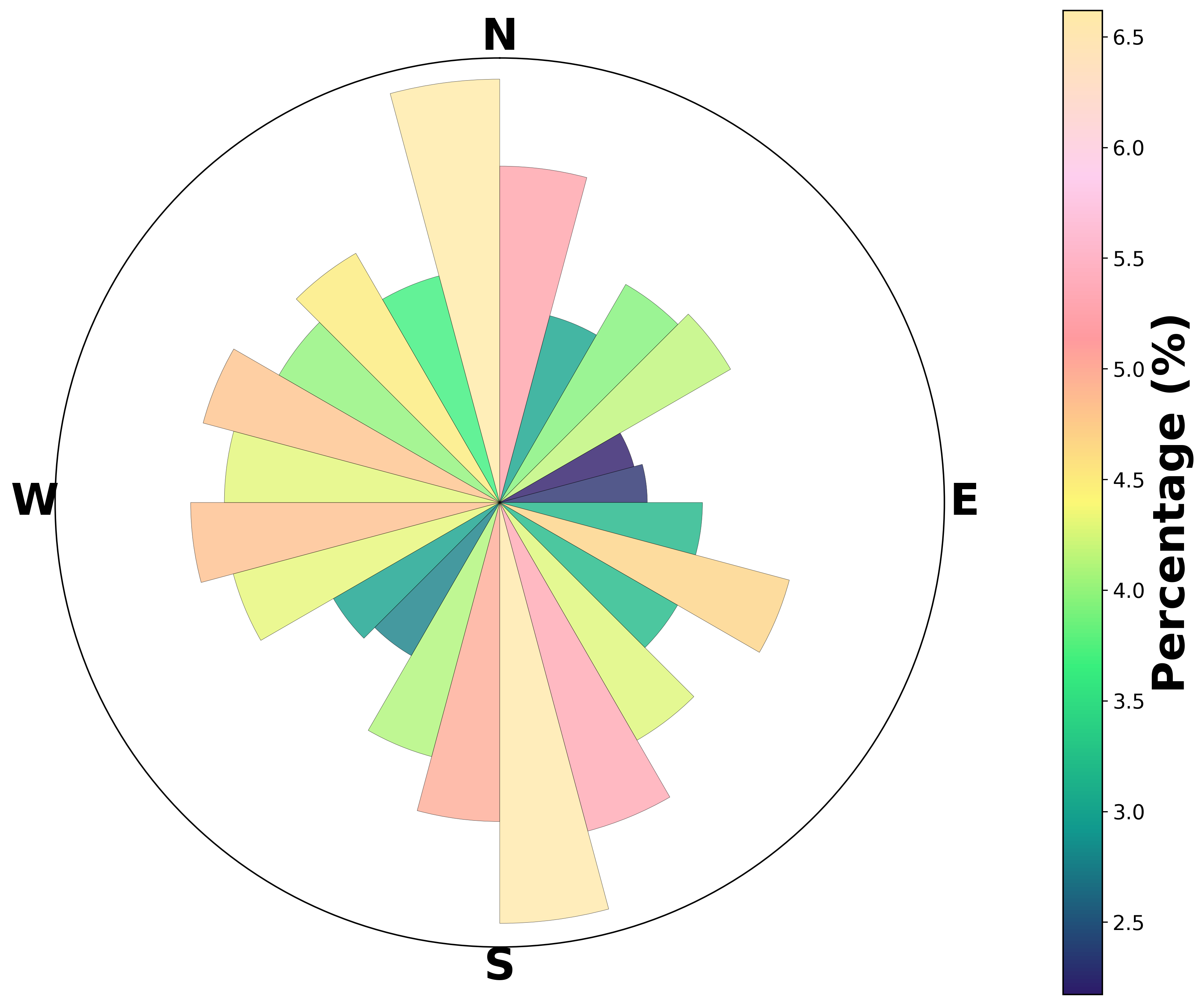}
        \caption{M$^3$CAD}
        \label{fig:m3cad_compass}
    \end{subfigure}
    \hfill
    \begin{subfigure}[b]{0.45\linewidth}
        \centering
        \includegraphics[width=\linewidth]{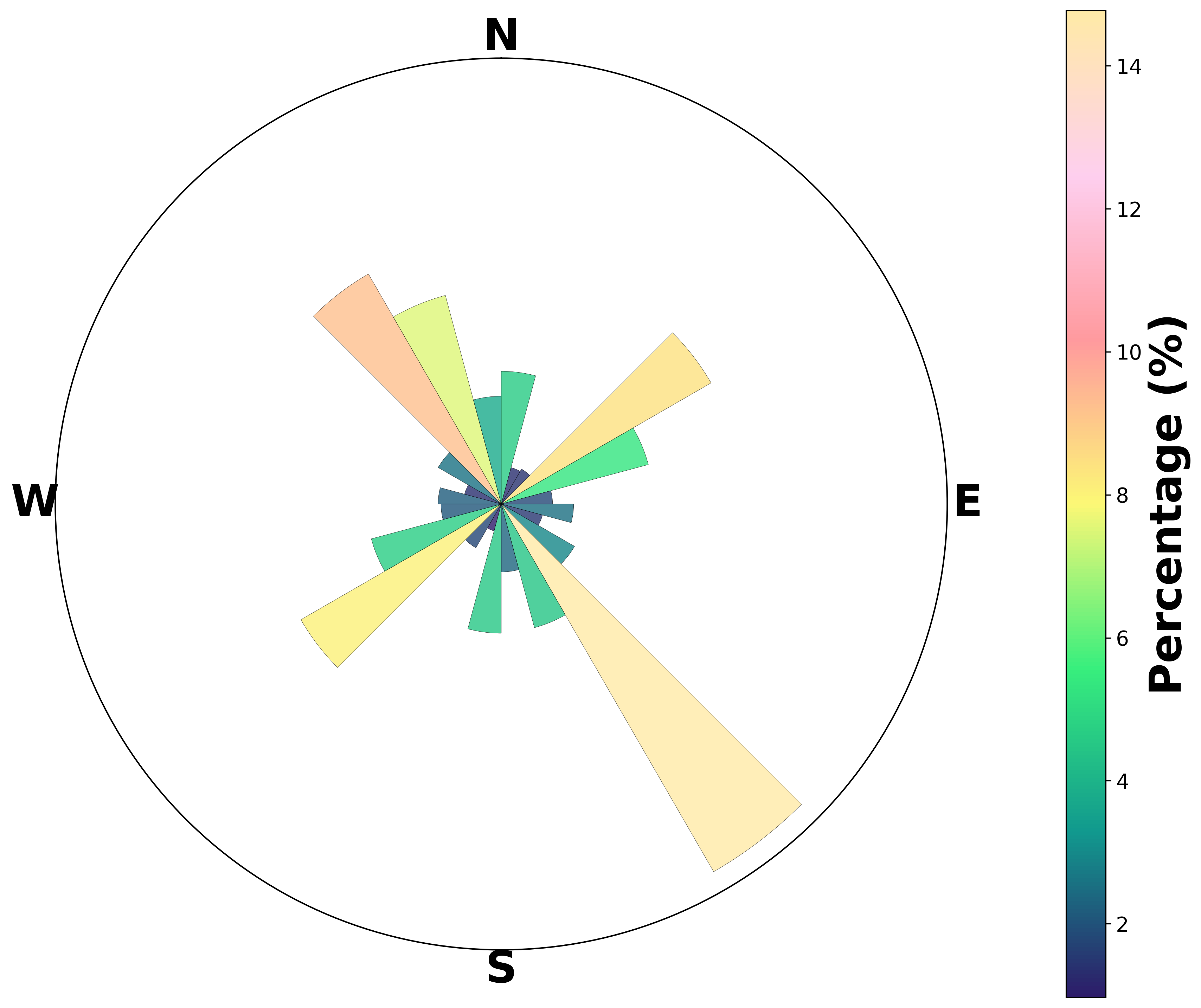}
        \caption{nuScenes}
        \label{fig:nuscene}
    \end{subfigure}
    \caption{\textbf{Directional distribution of ego-vehicle trajectories in M$^3$CAD and nuScenes.} The polar plots show the normalized distribution of ego-vehicle trajectory segments across movement directions, with wedge length and color intensity indicating the percentage of trajectories in each 15° angular bin. Trajectory counts are normalized across both datasets to ensure fair comparison. (a) M$^3$CAD exhibits a balanced distribution across directions, reflecting comprehensive scenario coverage in simulation. (b) nuScenes displays a strong concentration in specific directions, suggesting a dataset bias toward straight-driving behaviors with fewer turning maneuvers. The broader directional coverage in M$^3$CAD supports more diverse training conditions for autonomous driving models.}

    \label{fig:comparison}
\end{figure}
\subsection{Importance of Camera Data}
While the previous experiments demonstrate the robustness of cooperative perception under various noise conditions, an equally important question arises: how critical is environmental perception data for autonomous driving? 
Prior research~\cite{li2024ego, zhai2023admlp} suggest that when datasets mainly contain simple ego-vehicle trajectories, path planning can rely solely on internal states, e.g., velocity, acceleration, and steering angle, without using camera or LiDAR data.
This happens because existing real-world datasets, e.g., nuScenes, often include oversimplified trajectories, where the ego vehicle mostly drives in straight lines for easier data collection and synchronization. 
We argue that this conclusion is misleading, as real-world driving is far more complex.
As shown in Fig.~\ref{fig:comparison}, M$^3$CAD contains more diverse ego-vehicle trajectories, making sensor data essential for understanding the surrounding environment and achieving safe path planning.

We compare UniAD’s performance on M$^3$CAD with Ego-MLP~\cite{li2024ego}, a planning method that relies mainly on internal states, e.g., speed, acceleration, steering, rather than perception data from cameras and LiDAR. 
As shown in Table~\ref{tab:planning}, Ego-MLP performs similarly to UniAD on nuScenes dataset ($0.35 m$ vs. $0.46 m$ on average L2 error), but its performance drops sharply on M$^3$CAD dataset ($2.04 m$ vs. $0.46 m$ on average L2 error). 
This $4.4 \times$ gap highlights that when trajectories show realistic and complex behaviors such as lane changes, turns, and multi-vehicle interactions, effective planning depends heavily on rich perception data. 
These results confirm that M$^3$CAD better captures the complexity of real-world autonomous driving.

\begin{table}[h]
    \caption{Path planning results of different solutions on nuScenes and M$^3$CAD benchmarks. $^\dagger$ Results on nuScenes are reported from~\cite{li2024ego}.}
    \centering
    \resizebox{\linewidth}{!}{
    \begin{tabular}{c|c|cccc|cccc}
    \hline
     &  & \multicolumn{4}{c|}{\textbf{Avg. L2 ($m$) $\downarrow$}} & \multicolumn{4}{c}{\textbf{ Avg. Col. ($\%$) $\downarrow$}}\\
     \textbf{Benchmarks} & \textbf{Methods} & 1s & 2s & 3s & Avg & 1s & 2s & 3s & Avg \\
    \hline
    \multirow{2}{*}{nuScenes$^\dagger$} 
    & Ego-MLP & \textbf{0.15} & \textbf{0.32} & \textbf{0.59} & \textbf{0.35} & \textbf{0.00} & 0.27 & 0.85 & \textbf{0.37} \\
    & UniAD & 0.20 & 0.42 & 0.75 & 0.46 & 0.02 & \textbf{0.25} & \textbf{0.84} & \textbf{0.37} \\
    \hline
    \multirow{2}{*}{M$^3$CAD} 
    & Ego-MLP   & 1.02 & 2.04 & 3.06 & 2.04 & 0.00 & 0.09 & 0.30 & 0.13 \\
    & UniAD & \textbf{0.34} & \textbf{0.46} & \textbf{0.58} & \textbf{0.46} & \textbf{0.00} & \textbf{0.00} & \textbf{0.07} & \textbf{0.02} \\
    \hline
    \end{tabular}}
    \vspace{-2mm}
    \label{tab:planning}
\end{table}

\section{Conclusions}
\label{sec:conclusion}

%
In this paper, we present M$^3$CAD, the first comprehensive benchmark designed for cooperative autonomous driving that supports multi-vehicle, multi-task, and multi-modality evaluation. 
%
%
%
Building upon this benchmark, we further introduce a multi-level fusion method that adaptively balances communication efficiency and perception accuracy by supporting BEV feature fusion, query fusion, and reference point fusion.
Experiments demonstrate that our approach not only advances cooperative perception within simulation but also transfers effectively to real-world datasets such as nuScenes, achieving strong improvements in data-limited settings.
We believe the M$^3$CAD benchmark, together with the proposed multi-level fusion method and benchmark evaluations, will help drive further research in cooperative autonomous driving and support its transition into real-world applications.

\bibliographystyle{IEEEtran}
\bibliography{IEEEabrv,references}

@inproceedings{jiang2023vad,
  title={Vad: Vectorized scene representation for efficient autonomous driving},
  author={Jiang, Bo and Chen, Shaoyu and Xu, Qing and Liao, Bencheng and Chen, Jiajie and Zhou, Helong and Zhang, Qian and Liu, Wenyu and Huang, Chang and Wang, Xinggang},
  booktitle={Proceedings of the IEEE/CVF International Conference on Computer Vision},
  pages={8340--8350},
  year={2023}
}

@article{li2023learning,
  title={Learning for vehicle-to-vehicle cooperative perception under lossy communication},
  author={Li, Jinlong and Xu, Runsheng and Liu, Xinyu and Ma, Jin and Chi, Zicheng and Ma, Jiaqi and Yu, Hongkai},
  journal={IEEE Transactions on Intelligent Vehicles},
  volume={8},
  number={4},
  pages={2650--2660},
  year={2023},
  publisher={IEEE}
}

@article{hu2022where2comm,
  title={Where2comm: Communication-efficient collaborative perception via spatial confidence maps},
  author={Hu, Yue and Fang, Shaoheng and Lei, Zixing and Zhong, Yiqi and Chen, Siheng},
  journal={Advances in neural information processing systems},
  volume={35},
  pages={4874--4886},
  year={2022}
}

@article{chen2024end,
  title={End-to-end autonomous driving: Challenges and frontiers},
  author={Chen, Li and Wu, Penghao and Chitta, Kashyap and Jaeger, Bernhard and Geiger, Andreas and Li, Hongyang},
  journal={IEEE Transactions on Pattern Analysis and Machine Intelligence},
  year={2024},
  publisher={IEEE}
}

@inproceedings{sadat2020perceive,
  title={Perceive, predict, and plan: Safe motion planning through interpretable semantic representations},
  author={Sadat, Abbas and Casas, Sergio and Ren, Mengye and Wu, Xinyu and Dhawan, Pranaab and Urtasun, Raquel},
  booktitle={Computer Vision--ECCV 2020: 16th European Conference, Glasgow, UK, August 23--28, 2020, Proceedings, Part XXIII 16},
  pages={414--430},
  year={2020},
  organization={Springer}
}

@inproceedings{hu2022st,
  title={St-p3: End-to-end vision-based autonomous driving via spatial-temporal feature learning},
  author={Hu, Shengchao and Chen, Li and Wu, Penghao and Li, Hongyang and Yan, Junchi and Tao, Dacheng},
  booktitle={European Conference on Computer Vision},
  pages={533--549},
  year={2022},
  organization={Springer}
}

@inproceedings{casas2021mp3,
  title={Mp3: A unified model to map, perceive, predict and plan},
  author={Casas, Sergio and Sadat, Abbas and Urtasun, Raquel},
  booktitle={Proceedings of the IEEE/CVF Conference on Computer Vision and Pattern Recognition},
  pages={14403--14412},
  year={2021}
}

@inproceedings{karkus2023diffstack,
  title={Diffstack: A differentiable and modular control stack for autonomous vehicles},
  author={Karkus, Peter and Ivanovic, Boris and Mannor, Shie and Pavone, Marco},
  booktitle={Conference on robot learning},
  pages={2170--2180},
  year={2023},
  organization={PMLR}
}

@inproceedings{hong2024multi,
  title={Multi-agent collaborative perception via motion-aware robust communication network},
  author={Hong, Shixin and Liu, Yu and Li, Zhi and Li, Shaohui and He, You},
  booktitle={Proceedings of the IEEE/CVF Conference on Computer Vision and Pattern Recognition},
  pages={15301--15310},
  year={2024}
}

@inproceedings{zhang2024ermvp,
  title={Ermvp: Communication-efficient and collaboration-robust multi-vehicle perception in challenging environments},
  author={Zhang, Jingyu and Yang, Kun and Wang, Yilei and Wang, Hanqi and Sun, Peng and Song, Liang},
  booktitle={Proceedings of the IEEE/CVF Conference on Computer Vision and Pattern Recognition},
  pages={12575--12584},
  year={2024}
}

@inproceedings{li2024ego,
  title={Is ego status all you need for open-loop end-to-end autonomous driving?},
  author={Li, Zhiqi and Yu, Zhiding and Lan, Shiyi and Li, Jiahan and Kautz, Jan and Lu, Tong and Alvarez, Jose M},
  booktitle={Proceedings of the IEEE/CVF Conference on Computer Vision and Pattern Recognition},
  pages={14864--14873},
  year={2024}
}

@misc{zhai2023admlp,
      title={Rethinking the Open-Loop Evaluation of End-to-End Autonomous Driving in nuScenes}, 
      author={Jiang-Tian Zhai and Ze Feng and Jinhao Du and Yongqiang Mao and Jiang-Jiang Liu and Zichang Tan and Yifu Zhang and Xiaoqing Ye and Jingdong Wang},
      year={2023},
      eprint={2305.10430},
      archivePrefix={arXiv},
      primaryClass={cs.CV},
      url={https://arxiv.org/abs/2305.10430}, 
}

@article{qu2024head,
  title={HEAD: A Bandwidth-Efficient Cooperative Perception Approach for Heterogeneous Connected and Autonomous Vehicles},
  author={Qu, Deyuan and Chen, Qi and Zhu, Yongqi and Zhu, Yihao and Avedisov, Sergei S and Fu, Song and Yang, Qing},
  journal={arXiv preprint arXiv:2408.15428},
  year={2024}
}

@inproceedings{qu2024sicp,
  title={SiCP: Simultaneous Individual and Cooperative Perception for 3D Object Detection in Connected and Automated Vehicles},
  author={Qu, Deyuan and Chen, Qi and Bai, Tianyu and Lu, Hongsheng and Fan, Heng and Zhang, Hao and Fu, Song and Yang, Qing},
  booktitle={2024 IEEE/RSJ International Conference on Intelligent Robots and Systems (IROS)},
  pages={8905--8912},
  year={2024},
  organization={IEEE}
}

@inproceedings{wang2024driving,
  title={Driving into the future: Multiview visual forecasting and planning with world model for autonomous driving},
  author={Wang, Yuqi and He, Jiawei and Fan, Lue and Li, Hongxin and Chen, Yuntao and Zhang, Zhaoxiang},
  booktitle={Proceedings of the IEEE/CVF Conference on Computer Vision and Pattern Recognition},
  pages={14749--14759},
  year={2024}
}

@inproceedings{zimmer2024tumtraf,
  title={Tumtraf v2x cooperative perception dataset},
  author={Zimmer, Walter and Wardana, Gerhard Arya and Sritharan, Suren and Zhou, Xingcheng and Song, Rui and Knoll, Alois C},
  booktitle={Proceedings of the IEEE/CVF conference on computer vision and pattern recognition},
  pages={22668--22677},
  year={2024}
}

@inproceedings{dosovitskiy2017carla,
  title={CARLA: An open urban driving simulator},
  author={Dosovitskiy, Alexey and Ros, German and Codevilla, Felipe and Lopez, Antonio and Koltun, Vladlen},
  booktitle={Conference on robot learning},
  pages={1--16},
  year={2017},
  organization={PMLR}
}

@article{xu2022cobevt,
  title={CoBEVT: Cooperative bird's eye view semantic segmentation with sparse transformers},
  author={Xu, Runsheng and Tu, Zhengzhong and Xiang, Hao and Shao, Wei and Zhou, Bolei and Ma, Jiaqi},
  journal={arXiv preprint arXiv:2207.02202},
  year={2022}
}

@inproceedings{hu2023planning,
  title={Planning-oriented autonomous driving},
  author={Hu, Yihan and Yang, Jiazhi and Chen, Li and Li, Keyu and Sima, Chonghao and Zhu, Xizhou and Chai, Siqi and Du, Senyao and Lin, Tianwei and Wang, Wenhai and others},
  booktitle={Proceedings of the IEEE/CVF conference on computer vision and pattern recognition},
  pages={17853--17862},
  year={2023}
}

@inproceedings{caesar2020nuscenes,
  title={nuscenes: A multimodal dataset for autonomous driving},
  author={Caesar, Holger and Bankiti, Varun and Lang, Alex H and Vora, Sourabh and Liong, Venice Erin and Xu, Qiang and Krishnan, Anush and Pan, Yu and Baldan, Giancarlo and Beijbom, Oscar},
  booktitle={Proceedings of the IEEE/CVF conference on computer vision and pattern recognition},
  pages={11621--11631},
  year={2020}
}

@inproceedings{chen2019cooper,
  title={Cooper: Cooperative perception for connected autonomous vehicles based on 3d point clouds},
  author={Chen, Qi and Tang, Sihai and Yang, Qing and Fu, Song},
  booktitle={2019 IEEE 39th International Conference on Distributed Computing Systems (ICDCS)},
  pages={514--524},
  year={2019},
  organization={IEEE}
}

@inproceedings{fcooper,
author = {Chen, Qi and Ma, Xu and Tang, Sihai and Guo, Jingda and Yang, Qing and Fu, Song},
title = {F-Cooper: Feature Based Cooperative Perception for Autonomous Vehicle Edge Computing System Using 3D Point Clouds},
booktitle={2019 ACM/IEEE Symposium on Edge Computing (SEC)},
pages = {88–100},
}

@inproceedings{vnet,
  title={V2vnet: Vehicle-to-vehicle communication for joint perception and prediction},
  author={Wang, Tsun-Hsuan and Manivasagam, Sivabalan and Liang, Ming and Yang, Bin and Zeng, Wenyuan and Urtasun, Raquel},
  booktitle={European Conference on Computer Vision},
  pages={605--621},
  year={2020},
  organization={Springer}
}

@inproceedings{xu2022opv2v,
  title={Opv2v: An open benchmark dataset and fusion pipeline for perception with vehicle-to-vehicle communication},
  author={Xu, Runsheng and Xiang, Hao and Xia, Xin and Han, Xu and Li, Jinlong and Ma, Jiaqi},
  booktitle={2022 International Conference on Robotics and Automation (ICRA)},
  pages={2583--2589},
  year={2022},
  organization={IEEE}
}

@inproceedings{xu2022v2x,
  title={V2x-vit: Vehicle-to-everything cooperative perception with vision transformer},
  author={Xu, Runsheng and Xiang, Hao and Tu, Zhengzhong and Xia, Xin and Yang, Ming-Hsuan and Ma, Jiaqi},
  booktitle={European conference on computer vision},
  pages={107--124},
  year={2022},
  organization={Springer}
}

@inproceedings{lu2023robust,
  title={Robust collaborative 3d object detection in presence of pose errors},
  author={Lu, Yifan and Li, Quanhao and Liu, Baoan and Dianati, Mehrdad and Feng, Chen and Chen, Siheng and Wang, Yanfeng},
  booktitle={2023 IEEE International Conference on Robotics and Automation (ICRA)},
  pages={4812--4818},
  year={2023},
  organization={IEEE}
}

@article{kitti,
   title =  {{KITTI}-360: A Novel Dataset and Benchmarks for Urban Scene Understanding in 2D and 3D},
   author = {Yiyi Liao and Jun Xie and Andreas Geiger},
   journal = {arXiv preprint arXiv:2109.13410},
   year = {2021},
}

@inproceedings{xu2023v2v4real,
  title={V2v4real: A real-world large-scale dataset for vehicle-to-vehicle cooperative perception},
  author={Xu, Runsheng and Xia, Xin and Li, Jinlong and Li, Hanzhao and Zhang, Shuo and Tu, Zhengzhong and Meng, Zonglin and Xiang, Hao and Dong, Xiaoyu and Song, Rui and others},
  booktitle={Proceedings of the IEEE/CVF Conference on Computer Vision and Pattern Recognition},
  pages={13712--13722},
  year={2023}
}

@dataset{DAIR-V2X2021,
title={Vehicle-Infrastructure Collaborative Autonomous Driving: DAIR-V2X Dataset},
author={Institue for AI Industry Research (AIR), Tsinghua University},
website={http://air.tsinghua.edu.cn/dair-v2x},
year={2021}
}

@inproceedings{v2x-seq,
  title={V2x-seq: A large-scale sequential dataset for vehicle-infrastructure cooperative perception and forecasting},
  author={Yu, Haibao and Yang, Wenxian and Ruan, Hongzhi and Yang, Zhenwei and Tang, Yingjuan and Gao, Xu and Hao, Xin and Shi, Yifeng and Pan, Yifeng and Sun, Ning and others},
  booktitle={Proceedings of the IEEE/CVF Conference on Computer Vision and Pattern Recognition},
  pages={5486--5495},
  year={2023}
}

@ARTICLE{v2x-sim,
  author={Li, Yiming and Ma, Dekun and An, Ziyan and Wang, Zixun and Zhong, Yiqi and Chen, Siheng and Feng, Chen},
  journal={IEEE Robotics and Automation Letters}, 
  title={V2X-Sim: Multi-Agent Collaborative Perception Dataset and Benchmark for Autonomous Driving}, 
  year={2022},
  volume={7},
  number={4},
  pages={10914-10921},
  doi={10.1109/LRA.2022.3192802}}

@inproceedings{detr,
  title={End-to-end object detection with transformers},
  author={Carion, Nicolas and Massa, Francisco and Synnaeve, Gabriel and Usunier, Nicolas and Kirillov, Alexander and Zagoruyko, Sergey},
  booktitle={European conference on computer vision},
  pages={213--229},
  year={2020},
  organization={Springer}
}

@inproceedings{univ2x,
  title={End-to-end autonomous driving through v2x cooperation},
  author={Yu, Haibao and Yang, Wenxian and Zhong, Jiaru and Yang, Zhenwei and Fan, Siqi and Luo, Ping and Nie, Zaiqing},
  booktitle={Proceedings of the AAAI Conference on Artificial Intelligence},
  volume={39},
  number={9},
  pages={9598--9606},
  year={2025}
}

@InProceedings{v2x-real,
author="Xiang, Hao
and Zheng, Zhaoliang
and Xia, Xin
and Xu, Runsheng
and Gao, Letian
and Zhou, Zewei
and Han, Xu
and Ji, Xinkai
and Li, Mingxi
and Meng, Zonglin
and Jin, Li
and Lei, Mingyue
and Ma, Zhaoyang
and He, Zihang
and Ma, Haoxuan
and Yuan, Yunshuang
and Zhao, Yingqian
and Ma, Jiaqi",
editor="Leonardis, Ale{\v{s}}
and Ricci, Elisa
and Roth, Stefan
and Russakovsky, Olga
and Sattler, Torsten
and Varol, G{\"u}l",
title="V2X-Real: A Largs-Scale Dataset for Vehicle-to-Everything Cooperative Perception",
booktitle="Computer Vision -- ECCV 2024",
year="2025",
publisher="Springer Nature Switzerland",
address="Cham",
pages="455--470",
isbn="978-3-031-72943-0"
}

@article{coopdetr,
  title={CoopDETR: A unified cooperative perception framework for 3D detection via object query},
  author={Wang, Zhe and Xu, Shaocong and Zhuang, Xucai and Xu, Tongda and Wang, Yan and Liu, Jingjing and Chen, Yilun and Zhang, Ya-Qin},
  journal={arXiv preprint arXiv:2502.19313},
  year={2025}
}

@article{Huang2024V2X-R,
  title={V2X-R: Cooperative LiDAR-4D Radar Fusion for 3D Object Detection with Denoising Diffusion},
  author={Huang, Xun and Wang, Jinlong and Xia, Qiming and Chen, Siheng and Yang, Bisheng and Wang, Cheng and Wen, Chenglu},
  journal={arXiv preprint arXiv:2411.08402},
  year={2024}
}

@misc{wang2025whales,
      title={WHALES: A Multi-Agent Scheduling Dataset for Enhanced Cooperation in Autonomous Driving}, 
      author={Yinsong Wang and Siwei Chen and Ziyi Song and Sheng Zhou},
      year={2025},
      eprint={2411.13340},
      archivePrefix={arXiv},
      primaryClass={cs.CV},
      url={https://arxiv.org/abs/2411.13340}, 
}

@misc{zhang2025comtp,
      title={Co-MTP: A Cooperative Trajectory Prediction Framework with Multi-Temporal Fusion for Autonomous Driving}, 
      author={Xinyu Zhang and Zewei Zhou and Zhaoyi Wang and Yangjie Ji and Yanjun Huang and Hong Chen},
      year={2025},
      eprint={2502.16589},
      archivePrefix={arXiv},
      primaryClass={cs.LG},
      url={https://arxiv.org/abs/2502.16589}, 
}

@article{li2022bevformer,
  title={BEVFormer: Learning Bird’s-Eye-View Representation from Multi-Camera Images via Spatiotemporal Transformers},
  author={Li, Zhiqi and Wang, Wenhai and Li, Hongyang and Xie, Enze and Sima, Chonghao and Lu, Tong and Qiao, Yu and Dai, Jifeng},
  journal={arXiv preprint arXiv:2203.17270},
  year={2022}
}

@inproceedings{huang2024actformer,
  title={Actformer: Scalable collaborative perception via active queries},
  author={Huang, Suozhi and Zhang, Juexiao and Li, Yiming and Feng, Chen},
  booktitle={2024 IEEE International Conference on Robotics and Automation (ICRA)},
  pages={14716--14723},
  year={2024},
  organization={IEEE}
}

@inproceedings{zeng2022motr,
  title={Motr: End-to-end multiple-object tracking with transformer},
  author={Zeng, Fangao and Dong, Bin and Zhang, Yuang and Wang, Tiancai and Zhang, Xiangyu and Wei, Yichen},
  booktitle={European conference on computer vision},
  pages={659--675},
  year={2022},
  organization={Springer}
}

@INPROCEEDINGS{wang2024emiff,
  author={Wang, Zhe and Fan, Siqi and Huo, Xiaoliang and Xu, Tongda and Wang, Yan and Liu, Jingjing and Chen, Yilun and Zhang, Ya-Qin},
  booktitle={2024 IEEE International Conference on Robotics and Automation (ICRA)}, 
  title={EMIFF: Enhanced Multi-scale Image Feature Fusion for Vehicle-Infrastructure Cooperative 3D Object Detection}, 
  year={2024},
  volume={},
  number={},
  pages={16388-16394},
  doi={10.1109/ICRA57147.2024.10610545}}

@inproceedings{liu2022bevfusion,
  title={BEVFusion: Multi-Task Multi-Sensor Fusion with Unified Bird's-Eye View Representation},
  author={Liu, Zhijian and Tang, Haotian and Amini, Alexander and Yang, Xingyu and Mao, Huizi and Rus, Daniela and Han, Song},
  booktitle={IEEE International Conference on Robotics and Automation (ICRA)},
  year={2023}
}

@article{wang2025cmp,
    title={Cmp: Cooperative motion prediction with multi-agent communication},
    author={Wang, Zehao and Wang, Yuping and Wu, Zhuoyuan and Ma, Hengbo and Li, Zhaowei and Qiu, Hang and Li, Jiachen},
    journal={IEEE Robotics and Automation Letters},
    year={2025},
    publisher={IEEE}
}

\end{document}